\pdfoutput=1

\documentclass[11pt]{article}

\usepackage{acl}
\usepackage{times}
\usepackage{latexsym}
\usepackage{amsmath}
\usepackage{amsfonts}
\usepackage{svg}
\usepackage{multicol}
\usepackage{float}
\usepackage[T1]{fontenc}

\usepackage[utf8]{inputenc}

\usepackage{microtype}

\usepackage{multirow}
\usepackage{graphicx}
\usepackage{placeins}
\usepackage{booktabs,makecell,tabularx}

\makeatletter
\def\ScaleIfNeeded{%
  \ifdim\Gin@nat@width>\linewidth
    \linewidth
  \else
    \Gin@nat@width
  \fi
}
\makeatother

\let\oldincludegraphics\includegraphics
\renewcommand\includegraphics[2][]{%
  \oldincludegraphics[width=\ScaleIfNeeded]{#2}
}

\title{A Novel Perspective to Look At Attention: Bi-level Attention-based Explainable Topic Modeling for News Classification}

\author{Dairui Liu, Derek Greene \and Ruihai Dong \\
Insight Centre for Data Analytics, Dublin \\
School of Computer Science, University College Dublin, Ireland \\
\texttt{dairui.liu@ucdconnect.ie} \texttt{
\{derek.greene, ruihai.dong\}@ucd.ie} \\
}

\begin{document}
\maketitle
\begin{abstract}
Many recent deep learning-based solutions have adopted the attention mechanism in various tasks in the field of NLP. However, the inherent characteristics of deep learning models and the flexibility of the attention mechanism increase the models' complexity, thus leading to challenges in model explainability. To address this challenge,  we propose a novel practical framework by utilizing a two-tier attention architecture to decouple the complexity of explanation and the decision-making process. We apply it in the context of a news article classification task. The experiments on two large-scaled news corpora demonstrate that the proposed model can achieve competitive performance with many state-of-the-art alternatives and illustrate its appropriateness from an explainability perspective. We release the source code here\footnote{\url{https://github.com/Ruixinhua/BATM}}.
\end{abstract}

\section{Introduction}
The attention mechanism is one of the most important components in recent deep learning-based architectures in natural language processing (NLP). In the early stages of its development, the encoder-decoder models \cite{Bahdanau2015NeuralMT,xu2015show} often adopted an attention mechanism to improve the performance achieved by capturing different areas of the input sequence when generating an output in the decoding process to solve issues arising in encoding long-form inputs. Subsequently, researchers have applied the attention mechanism to large-scale corpora and developed a range of pre-trained language models \cite{kalyan2021ammus}, such as BERT \citep{devlin-etal-2019-bert} and GPT-1 \cite{radford2018improving}. This has yielded great progress across a range of NLP tasks, including sentiment analysis \cite{zhao2021bert} and news classification \cite{Wu2021FastformerAA}. However, the inherent characteristics of deep learning models and the flexibility of the attention mechanism increase these models' complexity, thus leading to challenges in model explainability.

Today, there is still no consensus among researchers regarding whether attention-based models are explainable in theory. Some researchers believe that attention weights may reflect the importance of features during the decision-making process and thus can provide an explanation of their operation if we visualize features according to their weight distribution \cite{luong-2015-effective, Lu2018topical}. However, other researchers have disagreed with this hypothesis. For example, Jain and Wallance's study demonstrated that learned attention weights are often uncorrelated with feature importance \cite{jain-2019-attention}. Some researchers have supported this viewpoint \cite{serrano-2019-attention}, but treated with skepticism by others  \cite{wiegreffe-2019-attention}.

In this paper, rather than validating the attention explainability theoretically, we propose a novel, practical explainable attention-based solution. Inspired by the idea of topic models \cite{blei2003latent}, our proposed solution decouples the complexity of explanation and the decision-making process by adopting two attention layers to capture \emph{topic-word} distribution and \emph{document-topic} distribution, respectively. Specifically, the first layer contains multiple attentions, and each attention is expected to focus on specific words from a topic. The attention in the second layer is then used to judge the importance of topics from the perspective of the target document. In order to further improve the model's explainability, we add an entropy constraint for each attention in the first layer. To prove the effectiveness of our proposed solution, we apply it in the context of a news article classification task and conduct experiments on two large-scaled news article datasets. The results presented later in Section \ref{sec:experiments} show that our model can achieve competitive performance with many state-of-the-art transformer-based models and pre-trained language models, while also demonstrating its appropriateness from an explainability perspective.


\section{Related Work}
\subsection{Attention Mechanism}
The attention mechanism was first applied on machine translation tasks \citep{Bahdanau2015NeuralMT} with the Seq2Seq model using RNN. To solve the dilemma in compressing long sequences by using an RNN-encoder, \citet{Bahdanau2015NeuralMT} introduced an attention mechanism by allowing RNN-decoder to assign attention weights to words in the input sequence. This strategy helps the decoder to effectively capture the relevant information between the hidden states of the encoder and the corresponding decoder's hidden state,  which avoids information loss and makes the decoder focus on the relevant position of the input sequence. This attention mechanism is named \textit{additive attention} or \textit{Tanh attention} because it uses the Tanh activation function. In our work, we propose to use additive attention to discover the underlying mixture of topics within a document. 

Furthermore, \citet{vaswani2017attention} proposed a transformer architecture to replace RNNs entirely with multi-head self-attention. This approach makes it possible to compute hidden representation for all input and output positions in parallel. The advantage of parallelized training has led to the emergence of many large pre-trained language models, such as BERT \citep{devlin-etal-2019-bert}. The improvement of using the transformer-based language model for generating representations is significant compared with popular word embedding methods such as GloVe \citep{pennington-etal-2014-glove}. However, along with the considerable enhancement in performance, it makes the attention-based language models difficult to interpret. One potential solution is to use attention weights to provide insights into the model.

\subsection{Attention as an Explanation}
The visualization of attention weight alignment in \citep{luong-2015-effective, vaswani2017attention} provides an intuitive explanation of the operation of additive attention and multi-head self-attention in machine translation tasks. But the faithfulness (i.e. accurately revealing the proper reasoning of the model) and plausibility (i.e. providing a convincing interpretation for humans) of using attention as an explanation for some tasks are still in debate, and the questioning is mainly on faithfulness \citep{jacovi-goldberg-2020-faithfulness}. This discussion is primarily focused on a simple model for specific tasks, such as text classification, using RNN models connecting an attention layer which is typically MLP-based \citep{Bahdanau2015NeuralMT}. A number of researchers have challenged the usefulness of attention as an explanation \citep{jain-2019-attention, serrano-2019-attention, bastings-2020-elephant}, concluding that saliency methods, such as gradient-based techniques, perform much better than using attention weights as interpretations in finding the most significant features of the input sequence that yield the predicted outcome. However, \citet{wiegreffe-2019-attention} claimed that, despite the fact that explanations provided by attention mechanisms are not always faithful, in practice, this does not invalidate the plausibility of using attention as an explanation.
We believe that the attention mechanism can provide a plausible explanation when applied correctly for an appropriate task.

\subsection{Role of Attention Mechanism}
Compared to simple additive attention, the Multi-Head Attention (MHA) mechanism, the core component of the big Transformer-based language model, is more complicated when attempting to interpret model behavior with complex weights distribution. Therefore, considerable work has attempted to understand the role played by the different attention heads \citep{Rogers2020BERTology}. For example, \citet{voita-etal-2019-MHSA} analyzed the patterns of attention heads by checking the survival of pruning, finding that the syntactic and positional heads are the final ones to be removed. \citet{kovaleva-etal-2019-revealing} identified five attention patterns of MHA, while \citet{Pande2021TheHH} proposed a standardized approach for analyzing patterns of different attention heads in the context of the BERT model. 

Instead of employing a complex transformer-like architecture with many MHA layers, we propose to start with a single MHA layer individually. Inspired by previous work, we focus on analyzing the role of attention heads in our architecture. We adopt a similar approach to \citep{Lu2018topical} by modeling attention using topics. However, unlike the topic attention model (TAN), which uses a bag-of-words (BOW) model based on variational inference to align the topic space and word space with extracting meaningful topics \citep{panwar-etal-2021-tan}, we assume that these multiple attention heads represent multiple topics in terms of their semantics. 

\begin{figure*}
    \centering
    \includegraphics{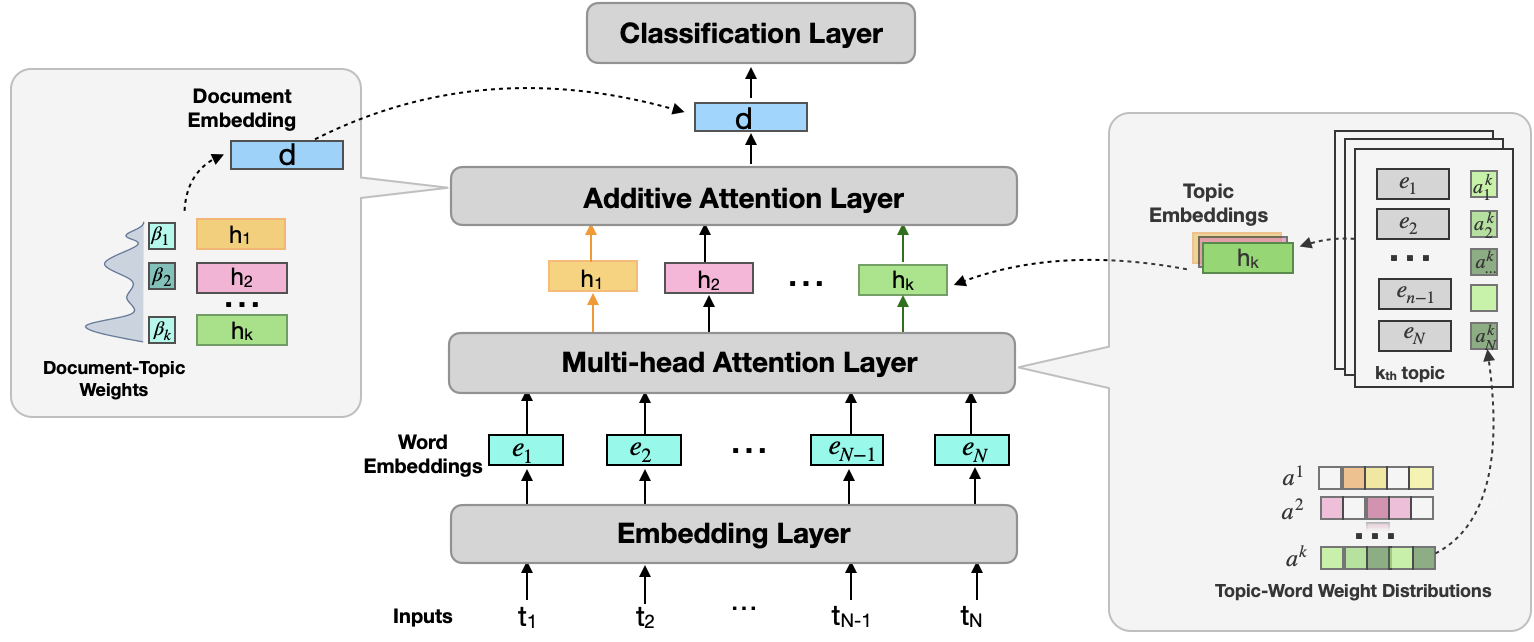}
    \caption{Structure of the proposed Bi-level Attention-based Topical Model (BATM).}
    \label{fig:BATM}
\end{figure*}

\section{Methodology}
\label{sec:methodology}
This section describes our proposed architecture Bi-level Attention-based Topical Model (BATM) as illustrated in Figure~\ref{fig:BATM}. It uses two attention layers to uncover a latent representation of the data and then makes use of attention weights as a form of topic distribution. We describe this architecture from the perspective of a news classification task.  
Our architecture consists of three components: an embedding layer, two attention layers, and a classification layer. After generating embedding vectors of words for the given news articles, we pass them to two attention layers to obtain the weight distribution of different words in each head (i.e. topic) and the weight distribution of different heads in the input articles. Then we generate the document representation vector based on these weights and finally classify the articles into different categories using a single linear layer. By analyzing the weight distribution of the attention layer on the entire news corpus, we find that some heads focus on the words related to the specific topics. These concentrated words help us understand the behavior of the attention mechanism.

\subsection{Embedding Layer}
There are two popular embedding methods: word-level embedding and contextual embedding, in general. Word-level embedding methods, such as GloVe, project different words into a word vector space and acquire a fixed-length word vector through a pre-trained embedding matrix. Contextual embedding models, such as BERT, generate different word vectors based on each word's context, so that the same word in different contexts can produce very different word vectors. For a given document $x$, suppose we have $N$ tokens in total, we use an appropriate tokenizer to partition it into tokens ${t_1, t_2, \ldots, t_N}$ according to the embedding method. Then we can represent the document using its embedding vectors ${e_1, e_2, \ldots, e_N}$ as an input to the attention layer.

\subsection{Multi-Head Attention Layer}
We use a multi-head attention mechanism to allow the model to focus on different positions in the document from different representation subspaces through multiple attention heads. We compute the weight distribution $g^{k}$ of the head vector $h_{k}$ through a single-layer feed-forward network first:
\begin{equation}
\mathrm{g}_{i}^{k}=v_{k} \tanh \left(\mathrm{W}_{k} e_{i}+ b_{k}\right)
\end{equation}
We then use the \textit{softmax} function to get the normalized weights distribution $\alpha^{k}$ among the document:
\begin{equation}
\alpha_{i}^{k}=\frac{e^{\mathrm{g}_{i}^{k}}}{\sum_{j}^{N} e^{\mathrm{g}_{j}^{k}}}
\end{equation}
Finally, the head vector $h_{k}$ is the weighted sum of word embedding vectors using the weights $\alpha^{k}$, given by
\begin{equation}
\label{formula:head_vector}
h_{k}=\sum_{i}^{N} \alpha_{i}^{k} e_{i}
\end{equation}
where trained parameters are $v_{k} \in \mathbb{R}^{D_{k}}, \mathrm{~W}_{k} \in \mathbb{R}^{E \times D_{k}}$, and ${b^{k}} \in \mathbb{R}^{D_{k}}$. $D_k$ is the projected dimension of each head in the middle, and $E$ is the embedding dimension, while the dimension of head vector $h_{k}$ is ${E}$ which is the same as embedding vector $e_{i}$ from Eqn.~\ref{formula:head_vector}.


\subsection{Additive Attention Layer}
For a given number of attention heads $K$, we have a group of head vectors $H=\left\{h_{1}, h_{2}, \ldots, h_{K}\right\}$, which are fed into an additive attention network to generate the document-topic distribution.
\begin{equation}
\begin{gathered}
\mu_{k}=c \tanh \left(\mathrm{W}_{H} h_{k}+b_{H}\right) \\
\beta_{k}=\frac{e^{\mu_{k}}}{\sum_{i}^{K} e^{\mu_{i}}} \\
\end{gathered}
\end{equation}
Finally, the document representation $d$ is the weighted sum of head vectors along with the weights distribution $\beta$ :
\begin{equation}
d=\sum_{i}^{K} \beta_{k} h_{k}  
\end{equation}
where trained parameters are $c \in \mathbb{R}^{D_{h}}, \mathrm{~W}_{H} \in \mathbb{R}^{E \times D_{h}}, b_{H} \in \mathbb{R}^{D_{h}}$, and the dimension of $d$ is also $E$ which is the same as $h_k$. 

\subsection{Classification Layer}
Since the representation of each document $d$ will be a dense vector containing a mixture of information about the document's content,  we can use it as the feature vector for the final news classification task:
\begin{equation}
y=softmax \left(\mathrm{W}_{C} d+b_{C}\right) \\
\end{equation}

\subsection{Entropy Constraint}
\label{sec:ec}
\begin{figure}
    \centering
    \includegraphics{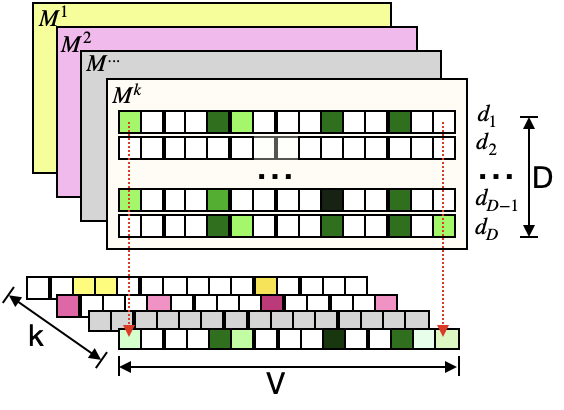}
    \caption{Structure of the topic-word weights $\alpha$ distribution among all documents.}
    \label{fig:weights_distribution}
\end{figure}

In order to further improve the explainability of our base model as described above, we now adjust the model so that each head only focuses on a specific set of words - i.e. we enforce topic-word weights distribution $\alpha^k$ not to spread over the document widely. We do this by computing the entropy of $\alpha^k$ as a part of the loss function. The entropy constraint penalizes the model when $\alpha^k$ has high entropy. Thus, the final loss with entropy constraint for the news classification task is:
\begin{equation}
\label{formula:entropy_constraint}
\mathcal{L}=\mathcal{L}_{C E}(y, \hat{y})+\lambda \frac{\sum_{k}^{K} \mathcal{E}_{\text {doc}}\left(\alpha^{k}\right)}{K} \\
\end{equation}
where $\mathcal{L}_{C E}(y, \hat{y})$ is the Cross-Entropy Loss between ground-truth class and predicted class, and $\lambda$ is a hyper-parameter to scale the magnitude of average entropy calculated by $\alpha^k$. The calculation for corresponding entropy $\mathcal{L}_{\text {entropy }}$ is by:
\begin{equation}
\label{formula:entropy_loss}
\mathcal{E}_{\text {doc}}\left(\alpha^{k}\right)=-\sum_{i}^{N} \alpha_{i}^{k} \log \alpha_{i}^{k}
\end{equation}
The entropy constraint applied on document-level in Eqn.~\ref{formula:entropy_loss} changes the distribution of topic-word weights $\alpha^{k}$. However, our goal is to find more diverse topics, which means different topics should focus on different words. Therefore, it is necessary to know how entropy decreases at the token level (i.e. across the vocabulary as shown in Figure~\ref{fig:weights_distribution}), which is defined by:
\begin{equation}
\label{formula:token_level}
\mathcal{E}_{\text {token}}\left(M_{i}\right)=-\sum_{k}^{K} M_{i}^{k} \log M_{i}^{k}
\end{equation}
To distinguish between the two variants of our model, we name the basic model as BATM-Base and use BATM-EC refer to the model with entropy constraints. From Eqn.~\ref{formula:entropy_constraint}, it is evident that if we set $\lambda$ as 0, BATM-EC will be equivalent to the basic model.

\subsection{Generating the Topic Distribution}
\label{subsec:gen_rep}

After training our proposed BATM model, we analyze the attention weights generated from the first attention layer (MHA) over the corpus vocabulary to generate a global topic distribution. Let us assume that there are $V$ words in the corpus and we have $K$ heads corresponding to $K$ topics. The resulting topic distribution takes the form of a $V \times K$ weight matrix, calculated from a trained MHA layer using embedded word vectors as inputs. Moreover, to identify the most important words for each topic (which we can view as being the topic's \emph{descriptor}), we extract the top-$T$ words from the topic distribution, which can help us understand the heads and interpret them as topics. We examine the interpretations of these topic descriptors and display some examples in Section~\ref{sec:evaluate_topic}.

\section{Experiments}
\label{sec:experiments}
We now evaluate the BATM model on two large-scale real-world datasets, and compare its performance with a number of state-of-the-art methods.

\subsection{Datasets}
We evaluate our proposed model on a news classification task and conduct extensive experiments on two public corpora. MIND \citep{wu-2020-mind} is a large-scale English dataset for news recommendation and categorization tasks. It contains information such as story title, abstract, and news category, but the public version does not include full article body content. We collected news articles from the Microsoft news website\footnote{We collect body content from \url{https://www.msn.com/en-ie/} using \url{https://github.com/msnews/MIND/tree/master/crawler}} to supplement it. There are 18 categories in the original MIND-large dataset, but three of them only have a small number of  articles ($<10$). Therefore, we exclude these categories from our experiment. The second one is the News Category Dataset\footnote{\url{https://www.kaggle.com/rmisra/news-category-dataset}} \cite{misra2018news,misra2021sculpting}, which contains approximately 200k news articles (each of them include a headline and a short news description) from 2012 to 2018 obtained from HuffPost. The original dataset has 41 categories, but some of these are duplicates. After merging the duplicated categories, there are 26 categories remain, which is denoted as News-26. 
We randomly split these two datasets into training/validation/test sets with a 80/10/10 split. Table~\ref{tab:statistic} summarizes the divisions and the key statistics of the datasets.

\begin{table*}
    \centering
    \begin{tabular}{lrrrrrr}
    \hline
    \thead{Dataset} & \thead{|Train|} & \thead{|Validation|}  & \thead{|Test|}  & \thead{Avg. Len} &  \thead{\#Class} & \thead{|Vocabulary|} \\
    \hline
    MIND-15 &   102,642 &  12,830 &   12,831 &  519.9 & 15 & 127,770 \\
    News-26 &   160,676 &  20,086 &   20,086 &  29.9 &  26 & 69,131 \\
    \hline
    \end{tabular}
    \caption{Statistical information for the MIND-15 and News-26 corpora. Note the vocabulary size only refers to English words without any punctuation or numbers.}
    \label{tab:statistic}
\end{table*}

\subsection{Baseline Models}
For the purpose of assessing classification performance, we first compare the effectiveness of our BATM base model relative to a number of attention-based and pre-trained language models:
\begin{itemize}
    \item BERT \citep{devlin-etal-2019-bert} composes of a bidirectional encoder of transformer and is pre-trained by using a combination of masked language modeling objective and next sentence prediction on a large corpus;
    \item DistilBERT \citep{Sanh2019DistilBERTAD} is a small, fast, cheap, and light transformer model trained by distilling BERT base;
    \item XLNet \citep{Yang2019XLNetGA} is an extension of the Transformer-XL \citep{dai-etal-2019-transformer} model, which utilizes an autoregressive method to learn bidirectional contexts by maximizing the expected likelihood over all permutations of input sequence factorization order;
    \item Roberta \citep{Liu2019RoBERTaAR} is a robustly optimized BERT that modifies key hyperparameters, removing the next-sentence pre-training objective and training with much larger mini-batches and learning rates;
    \item Longformer \citep{Beltagy2020LongformerTL} is based on RoBERTa \citep{Liu2019RoBERTaAR} and uses sliding window attention and global attention to model local and global contexts;
    \item Fastformer \citep{Wu2021FastformerAA} uses additive attention to perform multi-head attention, which is more efficient than a standard transformer.
\end{itemize}

The initial weights of these pre-trained language models (BERT, DistilBERT, XLNet, Roberta, and Longformer) are provided by Hugging Face Transformer \citep{wolf-etal-2020-transformers} library\footnote{The weights can download from the library:~\url{https://github.com/huggingface/transformers}}. We use a linear classifier to receive the pooled output from previous transformer layers and then fine-tune these models to adapt them to the classification task. For the attention-based model, Fastformer, we initialize its embedding matrix using GloVe embedding and follow the hyper-parameter settings in \citep{Wu2021FastformerAA}. 

\begin{figure*}
    \centering
    \includegraphics[width=\textwidth, height=1.5]{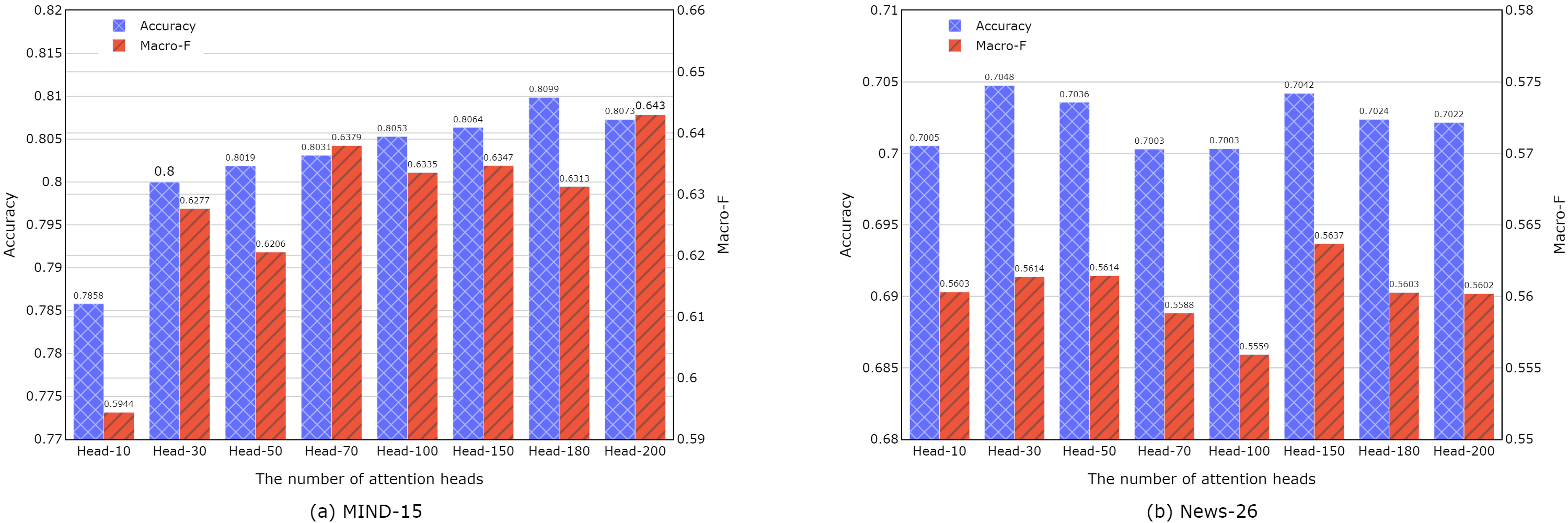}
    \caption{Performance of BATM-Base-GloVe with different number of attention heads on MIND-15 and News-26}
    \label{fig:head_test}
\end{figure*}

\subsection{Experimental Settings}
In our experiments, we consider two ways to initialize our embedding matrix: \emph{GloVe} embedding \citep{pennington-etal-2014-glove} and context embeddings from a pre-trained language model \emph{DistilBERT}~\cite{Sanh2019DistilBERTAD}, where embedding weights are not fixed during the training procedure. 
%
We examine how different number of heads would influence the performance of our proposed model on the validation set, the details is shown in Figure \ref{fig:head_test}. Unsurprisingly, on the MIND data set, the model needs to set a relatively larger number of topics, because the average length of news articles in the MIND dataset and its vocabulary size are much larger than the News-26 dataset, as indicated in Table \ref{tab:statistic}. We identify the number of topics for MIND-15 and News-26 as 180 and 30 for the rest of experiments, respectively. We use Adam \citep{Kingma2015AdamAM} for model optimization, and each epoch decays the learning rate by half.

\begin{table*}
\centering
    \begin{tabular}{lrrrr}
    \hline
    \multirow{2}{*}{\thead{Models}} & \multicolumn{2}{c}{\thead{MIND-15}} & \multicolumn{2}{c}{\thead{News-26}} \\
    \cline{2-5} 
     & \thead{Accuracy}  &    \thead{Macro-F} & \thead{Accuracy}  &   \thead{Macro-F} \\
    \hline
    BERT &       82.12±0.31 &  67.09±0.47 & 75.01±0.31 &  62.08±0.36 \\
    DistilBERT & 82.03±0.52 &  67.24±0.59 & 74.97±0.29 &  61.87±0.35 \\
    XLNet &      82.37±0.18 &  67.75±0.43 & 73.99±0.29 &   60.80±0.44 \\
    Roberta &    82.45±0.72 &  67.77±1.06 & 74.81±0.22 &  61.76±0.27 \\
    Longformer & 82.71±0.16 & 68.09±0.40   & 74.87±0.29 &  61.79±0.35 \\
    Fastformer-GloVe & 79.97±0.24 &  63.62±0.23 & 69.33±0.26 &  54.92±0.33 \\
    \hline
    BATM-Base-GloVe  &  79.75±0.15 &  63.24±0.41 & 69.72±0.16 & 55.53±0.12 \\
    BATM-Base-DB     & \textbf{82.82±0.15} &  \textbf{68.79±0.26} & \textbf{75.74±0.17} &  \textbf{63.01±0.23} \\ 
    \hline
    \end{tabular}
    \caption{Comparison of performance of models for the news classification task on MIND-15 and News-26 datasets. The best average scores are highlighted in bold.}
    \label{tab:mind_performance}
\end{table*}
\subsection{Performance Comparison}
The large pre-trained transformer variants perform better than the model with GloVe embedding, both for MIND-15 and News-26. Compared to Fastformer-GloVe, our BATM-Base-GloVe model achieves a similar result (variance in 0.3\% of accuracy and 0.4\% of Macro-F) for MIND-15 and a better result (variance in almost 0.4\% of accuracy and 0.6\% of Macro-F) for News-26. The differing results in MIND-15 and News-26 are due to the length of articles. As an efficient Fastformer can take a much longer sequence as input, it is advantageous to deal with long sequences which are unavailable in a short-length news dataset such as News-26. Using the pre-trained transformer-based embedding greatly improves the performance of our proposed BATM-Base model compared to the GloVe embedding, although it adds to the difficulty of interpretation. The performance difference of the other pre-trained language models with the BATM-Base-DB model is less than 1\% accuracy and approximately 2\% Macro-F, both for MIND-15 and News-26. These experiments demonstrate the effectiveness of our proposed model in constructing document representations. Thus, the analysis of BATM's behavior using the topic-word distribution and document-topic distribution is essential to understanding the role of Bi-level attention layers.

\begin{table*}[ht!]
    \centering
    \begin{tabularx}{\linewidth}{l X l }
        \toprule
        \thead{Label} & \thead{Topic Descriptor} & \thead{$C_{v}$} \\
        \midrule
        \multirow{4}{*}{Partisan} & indictments voter votes fiscal impeachment petitions electorate partisanship repudiation treasonous repeal majorities dissent amendments judicial electoral repealing elections ratification partisan incompetence conviction impeach justification resignations &  \multirow{4}{*}{0.76} \\
        \addlinespace[8pt]
        \multirow{3}{*}{Household} & cloth decorate towels embroidery basketballs suede bedding eggs fleece linen slippers cotton hooded porcelain bag plastic washed bowls clothes shirt flannel jacket jackets sweatshirt decorative &  \multirow{3}{*}{0.73} \\
        \midrule
        \multirow{4}{*}{Unknown} & serveware depositors mcadoo resold appliance cleats stockholders zoku horseshoes mailboxes frp hardwood holders multipacks disks unusable slugger noxzema laminate drawers tabletops ingvar costra memorabilia mailbox &  \multirow{4}{*}{0.61} \\
        \addlinespace[8pt]
        \multirow{4}{*}{Gender} & bisexuals affectional transpeople asexuals genderqueer cisgender queerness cisgendered discimination heterosexism courtyards bisexuality cissexism ochre asexuality sexualities heterosexuality androgyny transphobia heterosexual butches trans slurs blacks heterosexuals &  \multirow{4}{*}{0.57} \\
        \midrule
        \multirow{4}{*}{Diseases} & triceps mumps soundproofed measles immunodeficiency listeria stepfamilies brees pronated workouts bestival talaq coronavirus stepfamily babyproofing salmonellosis obliterans varicella homestyle iguodala bomer griever botulism gbk cortisol &  \multirow{3}{*}{0.45} \\
        \addlinespace[8pt]
        \multirow{3}{*}{Schedule} & said evening keynote annual month morning event scheduled weekend attended week according adjusted hosted inaugural host conferences conference attend telecast afternoon night will brightness sessions &  \multirow{3}{*}{0.38} \\
        \bottomrule
    \end{tabularx}
    \caption{Examples of topics identified by our approach, in terms of extracted topic descriptors, topic coherence scores $C_{v}$, and manually-assigned labels.}
    \label{tab:topic_words}
\end{table*}

\begin{table*}
    \centering
    \begin{tabular}{lrrrrrrrr}
    \hline
    \multirow{2}{*}{\thead{$\lambda$}} & \multicolumn{4}{c}{\thead{MIND-15}} & \multicolumn{4}{c}{\thead{News-26}} \\
    \cline{2-9} 
     &  \thead{Accuracy$ \uparrow$} &  \thead{Macro-F$ \uparrow$} &  \thead{Avg.$ \mathcal{E}_{doc} \downarrow$} &  \thead{Avg.$\mathcal{E}_{token} \downarrow$} &  \thead{Accuracy$ \uparrow$} &  \thead{Macro-F$ \uparrow$} &  \thead{Avg.$\mathcal{E}_{doc} \downarrow$} &  \thead{Avg.$\mathcal{E}_{token} \downarrow$}\\
    \hline
    0    &      80.50 &    63.40 &     3.171 &      8.542 &     70.03 &   55.88 &        2.175 &          9.022 \\
    1e-6 &     80.13 &    62.97 &    3.049 &      8.483 &     69.43 &   54.96 &        2.176 &          9.073 \\
    1e-5 &     80.16 &    64.07 &    3.076 &      8.599 &     69.55 &   55.12 &        2.129 &          8.995 \\
    1e-4 &     79.03 &    61.35 &    2.251 &      7.624 &     69.39 &   54.74 &        1.943 &          8.879 \\
    1e-3 &     72.86 &    50.58 &    0.041 &      5.947 &     58.16 &   38.74 &        0.080 &          7.071 \\
    1e-2 &     65.66 &    36.39 &    0.002 &      4.464 &     49.36 &   27.79 &        0.009 &          7.355 \\
    \hline
    \end{tabular}
    \caption{Influence of $\lambda$ of BATM-EC model on MIND-15 and News-26 datasets with 180 and 30 heads respectively.}
    \label{tab:entropy_factor}
\end{table*}

\subsection{Evaluation of Global Topic Representation}
\label{sec:evaluate_topic}
Besides the classification performance, we are also interested in whether each extracted topic descriptor as described in \ref{subsec:gen_rep} has an intuitive meaning. We take the top-25 highest scoring terms from each topic and calculate topic coherence scores $C_{v}$ \citep{Rder2015Coherence}. The average coherence scores of all topics of the \emph{BATM-Base-GloVe} model are 0.58 and 0.56 on the MIND dataset and the news category datasets, respectively. Moreover, to more intuitively understand the meaning of topics mined by our model, we list a few topic examples whose coherence scores range from 0.3 to 0.8 along with a manually-assigned label in Table \ref{tab:topic_words}. The topics with coherence scores between 0.55 and 0.8 usually have precise meanings, such as the topic labeled as ``Partisan" score of 0.76, where the vast majority of words are related to political activities and elections. However, some topics with a score in the range of $0.55\sim0.8$ are still tough to surmise the focus, as the unknown topic (labeled as ``Unknown" with $C_{v}$ value is 0.61) suggest, where the correlation of topic descriptors is non-intuitive.
In contrast, some low-coherence topics may contain highly relevant words as well. For example, the topic ``Schedule" with a score of 0.38 (under 0.55) mainly includes words related to time and arrangement, which we can comprehend the central point of these words, but the automated metric unfairly evaluates it. Therefore, with the auxiliary of topic coherence measurement and manual verification, we are firmly convinced that topic descriptors extracted by the \emph{BATM-Base-GloVe} model indeed have specific meanings.

\section{Effect of Entropy Constraints}

In the previous sections, the proposed BATM-Base-GloVe model demonstrates its competitive classification performance and excellent explainability. We now study the effect of adding an entropy constraint, as discussed in Section~ \ref{sec:ec}. In the extended model, referred to as BATM-EC, $\lambda$ determines the degree of constraint that is imposed, so the BATM-Base-GloVe model is a special case when $\lambda$ is zero.

This study assumes that a good topic (a first-level of attention) should only focus on specific words related to that topic. Its weight distribution on a news article should not be flat for the whole document, while its global weight distribution should also not be widely spread out across the entire vocabulary (i.e., it should have a relatively lower entropy ). Therefore, we observe the dynamic of two entropy metrics \emph{$\mathcal{E}_{doc}$} and \emph{$\mathcal{E}_{token}$} (see calculation in Eqn.~\ref{formula:entropy_loss} and Eqn.~\ref{formula:token_level}) by setting different values of $\lambda$. We present the performance and entropy changes along with the values of $\lambda$ in Table~\ref{tab:entropy_factor}

The results meet our expectations. When $\lambda$ reaches le-4, both entropy indicators decrease significantly with an acceptable trade-off in classification performance. When continually increasing the impact of entropy constraints, both entropy indicators and classification performance decrease dramatically. This is reasonable, as this experiment is conducted with a fixed number of heads. When attention focuses on a minimal number of topics, and the number of topics does not increase accordingly, information within article texts is likely to be lost, affecting the classification performance.

\section{Discussion and Future Work}
While the variant of our proposed model, BATM-base-DB, which is initialized by the contextual embeddings, can outperform all alternatives, the meaning of its topics is much worse than BATM-Base-GloVe. Each contextual embedding learned by pre-trained language models will merge the information from its surrounding words, which increases the difficulties of the proposed attention layer to capture the topics it focuses on, thus leading to more noise in their representations. 

Another challenge we will address in the future is how to balance the computation cost, topic granularity, and classification performances. As discussed in the previous sections, it will affect the model's classification performance if we only introduce entropy constraints without incrementing the number of attention heads. However, increasing the number of attention heads will lead to the proportional increment of parameters, increasing the complexity of the model and resulting in a high computation cost. We will consider increasing the number of heads and the extending entropy constraint further, to improve classification performance while maintaining strong explainability. 



\section{Conclusion}
In this paper, we presented a novel approach that harnesses a bi-level attention framework to decouple the text classification process as topic capturing, topic importance recognition and decision-making process to benefit explainability. We conducted the experiments on two large-scale text corpora. Compared with a number of state-of-the-art alternatives on a text classification task, our model can not only achieve a competitive performance, but also demonstrates a strong ability to capture intuitive meanings in the form of topical features, thus improving its explainability and transparency. In addition, by initializing it with contextual embeddings, our model outperforms all the baseline models. 

\vskip 0.6em
\noindent\textbf{Acknowledgements.} This research was supported by Science Foundation Ireland (SFI) under Grant Number SFI/12/RC/2289\_P2.

\bibliography{anthology,reference}
\bibliographystyle{acl_natbib}

\appendix
\section{Appendix}
\label{sec:appendix}
\subsection{Experimental Environment}
Our experiments are conducted on the sonic system with Linux operating system. We use PyTorch 1.8.0 as the backend. The GPU type is Nvidia Tesla V100 and A100 with 32GB and 40GB GPU memory, respectively. We run each experiment 5 times with fixed random seeds by a single thread.
\subsection{Preprocessing}
We use the PyTorch default Tokenizer to preprocess texts. And we remove all the non-alphabetic characters when extracting the topic descriptors from the first attention layer.
\subsection{Hyperparameter Settings}
The dimension of the GloVe embedding and pre-trained language model (PLM) is 300 and 768, respectively. 
The learning rate for the GloVe-based model and PLM model is 1e-3 and 5e-5, respectively. The maximum sequence length of all models is 512 on MIND-15 and 100 on News-26, except for Fastformer, which is 2048 on MIND-15.
The batch size is 32 for all experiments, both on MIND-15 and News-26. 
\subsection{More Topic Examples}
See Table \ref{tab:appendix_topic}.
\newpage
\begin{table*}
    \centering
    \begin{tabularx}{\linewidth}{l X l }
        \toprule
        \thead{Label} & \thead{Topic Descriptor} & \thead{$C_{v}$} \\
        \midrule
        \multirow{4}{*}{Gender} & lgbtq divorce lgbt infertility divorced hiv transgenders stepparent hpv surrogacy divorcing honeymoons heterosexuals marriage honeymoon transgendered weddings prenuptial menopause premarital alimony stepfamily listeria queer prenups&  \multirow{4}{*}{0.71} \\
        
        \addlinespace[10pt]
        \multirow{3}{*}{Mood} & disabling rebooting attacker accidently alerted viewer snapshots reset incriminating device disables inadvertently maliciously alerting securely jagged unintentionally sobering unsettling crashing gruesome wreckage jarring helpfully accidentally &  \multirow{3}{*}{0.70} \\
        \midrule
        \multirow{3}{*}{Marriage} & bridal wedding playdates preschooler brides toddlers bride gradeschool kindergarten mehndi kids toddler carolee weddings pacifier uighur preschoolers kyiv boomer udaipur design bridesmaid kid preschool kindergarden &  \multirow{3}{*}{0.67} \\
         \addlinespace[10pt] 
         \multirow{3}{*}{Disease} &epidemiology smashbox hilson dietetics nondairy deminers ijustine kimmel circadian vitamix presenteeism disinformers preparers disick keri fearless jwt integrative fassbender engelberg nutritionists swizz nivea juanes braff &  \multirow{3}{*}{0.67} \\
         \midrule
         \multirow{3}{*}{Unknown} & succinct republished talkbacks commenter peterman compiling errico excerpted newsfeeds reposted techdirt compiled dealnews compiles emailer tipsters editors crossposted postings downloaded collated tipster rnberg snarkiest khayr & \multirow{3}{*}{0.53} \\ 
         \addlinespace[10pt]
         \multirow{3}{*}{Law} & larceny forgery summonses unlawful misstatements offences felonies indictments wrongdoing audits contemplated misconduct misstatement breach burglary perjury incidents defendants tolerances irregularities misdemeanor fabricated misdemeanors comply statutory
         & \multirow{3}{*}{0.5} \\ 
         \midrule
         \multirow{3}{*}{Relationship} & son playgroup aged daughter womb nieces playdate granddaughters mums playroom parents ladera swingset sons playdates picnicking tykes toddlers icmi eldest napped dad newborn children bedtimes & \multirow{3}{*}{0.48} \\ 
         \addlinespace[10pt] 
         \multirow{4}{*}{Unknown} & workarounds reposting malicious voicemails emboldening excerpted harpersbazaar screenshots mischaracterizing defamatory incriminating formatted manipulates maliciously repost screenshot keystrokes enraging downloaded fallible poignant undeleted snapshots overwritten succinct & \multirow{4}{*}{0.42} \\ 
        \midrule
        \multirow{3}{*}{Sports} & women bicycle home races bike boats racing run wheelchair floors wife walking Minnesota race rentals volleyball Tennessee girls couples basketball clubs flying cars beach golf &  \multirow{3}{*}{0.33} \\
        \addlinespace[10pt] 
        \multirow{3}{*}{Unknown} & bellefonte balcones ellijay intracoastal titusville asbury masterson kander riverhead hallandale whidbey bridgehampton hiawatha bedminster boylston rossville schertz bushnell chaska rayden riverdale boothbay simcoe deerfield millcreek & \multirow{3}{*}{0.3} \\ 
        \bottomrule
    \end{tabularx}
    \caption{Examples of topics identified by our approach, in terms of extracted topic descriptors, topic coherence scores $C_{v}$, and manually-assigned labels.}
    \label{tab:appendix_topic}
\end{table*}

\end{document}